\documentclass[journal,hideappendix]{vgtc}        

\onlineid{0}

\vgtccategory{Research}

\vgtcpapertype{algorithm/technique}

\newcommand{\sysname}{ConfEviSurrogate} 

\title{{\sysname}: A Conformalized Evidential Surrogate Model for Uncertainty Quantification}

\author{%
  \authororcid{Yuhan Duan}{0009-0006-8101-3488},
  Xin Zhao,
  Neng Shi,
  and Han-Wei Shen
}

\authorfooter{
  \item
  	Yuhan Duan, Xin Zhao, Neng Shi, and Han-Wei Shen are with The Ohio State University.
  	E-mail: \{duan.418\,$|$\,zhao.2892\,$|$\,shi.1337\,$|$\,shen.94\}@osu.edu\,

}

\abstract{%
Surrogate models, crucial for approximating complex simulation data across sciences, inherently carry uncertainties that range from simulation noise to model prediction errors. Without rigorous uncertainty quantification, predictions become unreliable and hence hinder analysis.
While methods like Monte Carlo dropout and ensemble models exist, they are often costly, fail to isolate uncertainty types, and lack guaranteed coverage in prediction intervals. 
To address this, we introduce {\sysname}, a novel Conformalized Evidential Surrogate Model that can efficiently learn high-order evidential distributions, directly predict simulation outcomes, separate uncertainty sources, and provide prediction intervals.
A conformal prediction-based calibration step further enhances interval reliability to ensure coverage and improve efficiency.
Our {\sysname} demonstrates accurate predictions and robust uncertainty estimates in diverse simulations, including cosmology, ocean dynamics, and fluid dynamics.

}


\graphicspath{{figs/}{figures/}{pictures/}{images/}{./}} 

\usepackage{tabu}                      
\usepackage{booktabs}                  
\usepackage{lipsum}                    
\usepackage{mwe}  
\usepackage{graphicx}

\usepackage{mathptmx}                  

\usepackage{amsfonts}
\usepackage{multirow}
\usepackage{makecell}
\usepackage{amsmath}
\usepackage{fontawesome}
\usepackage{algorithm}
\usepackage{amsthm}
\newtheorem{theorem}{Theorem}

\usepackage[noend]{algpseudocode}

\begin{document}



\maketitle
\section{Introduction}

Ensemble simulations are widely used in scientific disciplines, including cosmology and hydrodynamics, to model complex natural phenomena~\cite{almgren2013nyx, ringler2013mpas,mallinson2013cloverleaf,gaudin2014optimising}.
These simulations typically involve solving systems of ordinary and partial differential equations with approximated numerical solutions.
%
However, high-fidelity simulations, especially those with high dimensionality, are computationally intensive and time-consuming, thus severely affecting the possibility of large-scale or interactive visualization and parameter explorations.
%
To overcome these computational challenges, machine learning (ML)-based surrogate models have emerged as efficient alternatives to traditional numerical simulations. 
Surrogate models can significantly accelerate data generation and facilitate rapid exploration of parameter space~\cite{han2022coordnet,hazarika2019nnva, he2019insitunet,shen2024surroflow,shi2022gnn,shi2022vdl}. 
%

Despite their efficiency, existing surrogate models still face several critical limitations. 
First, many models lack robust and efficient uncertainty quantification (UQ)~\cite{kamal2021recent, molnar2024uncertainty}. 
They typically learn deterministic one-to-one mappings between simulation parameters and output, ignoring the need to model uncertainty.
Ensemble datasets used for training are approximations of real phenomena, and the surrogate models built upon these datasets introduce an additional layer of approximation.
%
%
%
%
As a result, both the data sets and the surrogate models inherently contain uncertainties, making it crucial to convey these uncertainties associated with the models. 
Although some surrogate methods provide UQ, such as dropout-based approaches~\cite{srivastava2014dropout, gal2016dropout} and ensemble-based approaches~\cite{lakshminarayanan2017simple}, they require multiple inference runs, leading to high computational costs. This contradicts the core goal of surrogate models: rapidly generating data for efficient exploration and visualization. 
%
%
Therefore, efficiently and accurately quantifying uncertainty remains a critical challenge.

%

Second, existing surrogate models often struggle to distinguish different sources of uncertainty, whether arising from the ensemble data or from the surrogate model’s prediction errors. This limitation significantly reduces the practical usefulness of surrogate models. Identifying uncertainty in the data can help scientists spot areas where data is inherently unstable or noisy, guiding them to improve their simulations effectively~\cite{zeng2022adaptive}. On the other hand, recognizing uncertainty from the surrogate model allows scientists to pinpoint where the surrogate model isn't reliable enough, thus increasing user confidence in the model. Therefore, clearly separating these two types of uncertainty is crucial for effectively evaluating both simulation quality and model reliability.
Third, current surrogate models typically lack interval-based predictions with coverage guarantees~\cite{molnar2024uncertainty}.
Coverage intervals ensure that the true simulation results lie within these predicted intervals at a user-specified probability. Without formal coverage guarantees, users lack confidence in these bands, limiting the practical utility and trustworthiness of the surrogate predictions.

To address these challenges, we propose {\sysname}, a novel architecture that combines evidential deep regression (EDR)~\cite{amini2020deep} and conformal prediction~\cite{vovk2005algorithmic, romano2019conformalized, shafer2008tutorial, angelopoulos2021gentle} for a robust and well-calibrated UQ. 
Specifically, our model learns a flexible evidential distribution of the simulation outputs to tackle the problem of lacking comprehensive UQ. The evidential distribution enables accurate simulation outcome predictions, clearly distinguishing between aleatoric (data-induced) and epistemic (model-induced) uncertainty, and producing reliable predictive intervals of simulation outcomes. 
To further ensure the reliability of these intervals, we incorporate a conformal prediction-based calibration step. This step adjusts the initial intervals generated by EDR, guaranteeing that the true outputs fall within the calibrated intervals at a user-specified coverage. While strict coverage guarantees often imply wider intervals, our approach can yield narrower sets without sacrificing the specified coverage level.
In addition, we develop an interactive visual interface that facilitates real-time exploration of simulation outputs and their associated uncertainties.

Our {\sysname} consists of three primary components. The first component is the EDR-based distribution modeling. 
We assume that our simulation data are sampled from a higher-order Student-t distribution. 
This distribution modeling allows the model not only to accurately predict simulation outcomes but also to simultaneously capture two key sources of uncertainty: (1) the inherent uncertainty in the simulation data, called aleatoric uncertainty, and (2) the predicted error of the surrogate model, called epistemic uncertainty. 
We utilize a convolutional neural network (CNN)-based architecture to learn the hyperparameters of this distribution, thereby deriving outcome predictions alongside the two types of uncertainty. We can also construct initial predictive intervals from the learned distribution, though these intervals lack formal theoretical coverage guarantees. 
To mitigate the lack of coverage guarantees, the second component applies conformal prediction for calibration, ensuring reliable and well-calibrated predictive intervals.
We integrate a model-agnostic conformal prediction method to calibrate these intervals, ensuring rigorous coverage guarantees.
Specifically, we use additional hold-out calibration data to compute a non-conformity score for each calibration sample, quantifying its prediction errors. 
Since the calibration dataset is drawn from the same distribution as the true data, these scores form an empirical distribution that reflects the model's real-world error behavior. 
We then leverage these scores to systematically adjust our initial intervals, ensuring that the bands reliably contain the true outcomes at a user-specified coverage level.
The third component of our surrogate is interactive visualization. We develop a visual interface that allows scientists to explore simulation results and analyze uncertainty estimates.

We demonstrate the effectiveness of the proposed approach in cosmology, fluid, and ocean simulations, and compare our model's predicted outputs with ground truth and existing surrogate models to assess accuracy and uncertainty estimation. In summary, the main contributions of this paper are:
\begin{itemize}
\item We introduce an evidential deep regression-based surrogate model that accurately predicts simulation outputs, robustly separates both epistemic and aleatoric uncertainties, and provides intuitive predictive intervals.
\item We incorporate a conformal prediction-based calibration method to produce narrower intervals with rigorous theoretical coverage guarantees, thereby enhancing both reliability and precision.
\item We employ an interactive visual interface enabling scientists to intuitively explore simulation outputs, uncertainty distributions, and coverage intervals, facilitating informed decision-making and deeper insights into simulation phenomena.
\end{itemize}

\section{Related Works}
In this section, we review related work in surrogate models for ensemble simulations, with a particular focus on models that incorporate uncertainty quantification.

\textit{Surrogate Models for Ensemble Data.}
In scientific computing, high-fidelity simulations are often computationally intensive. To accelerate these processes, significant research has focused on developing surrogate models as efficient alternatives.
Early work by Bhatnagar et al.~\cite{bhatnagar2019prediction} trains a convolutional neural network to predict steady 2D flow fields (pressure and velocity) around airfoils, achieving results orders of magnitude faster than a Reynolds-averaged Navier–Stokes solver. 
Meanwhile, He et al.~\cite{he2019insitunet} propose InsituNet, a generative adversarial network-based surrogate model that produces accurate image-based predictions directly from simulation parameters.
To facilitate user interactions and exploration, Li et al.~\cite{li2024paramsdrag} introduce ParamsDrag, enabling intuitive manipulation of simulation outcomes.
To handle the irregular mesh structures commonly found in scientific datasets, Shi et al.~\cite{shi2022gnn} introduce GNN-surrogate, which effectively leverages graph neural networks. For high-resolution ensemble visualization, Shi et al.~\cite{shi2022vdl} also propose VDL-Surrogate, a view-dependent deep surrogate model that uses latent representations to address memory constraints.
To capture multi-scale physics, Le and Ooi~\cite{le2021surrogate} design a multigrid-inspired U-Net surrogate that embeds coarse-to-fine feature hierarchies.
Together, these works demonstrate the growing adoption of diverse machine learning architectures to tackle various challenges in scientific visualization, such as improving predictive accuracy, handling unstructured data, and mitigating memory limitations.  
However, most of these models do not quantify uncertainty, which is critical for robust scientific decision-making and risk assessment.

\textit{Surrogate Models for Uncertainty Quantification.}
Uncertainty quantification (UQ) plays a central role in surrogate modeling by assessing prediction reliability and guiding downstream decisions. In recent years, increasing attention has been devoted to developing UQ-capable surrogate models.
Xu et al.~\cite{xu2022using} employ a polynomial chaos expansion (PCE)-based surrogate to emulate runoff generation in an Earth system model. PCE can capture complex, nonlinear input–output relationships and supports global sensitivity analysis.
Schram et al.~\cite{schram2023uncertainty} review common approaches to UQ in data-driven surrogates, including deep ensembles, Bayesian neural networks, and quantile regression.
Shen et al.~\cite{shen2024surroflow} propose a normalizing flow-based surrogate model to learn complex output distributions in scientific simulations. To handle resolution-induced uncertainty, they also introduce PSRFlow~\cite{shen2023psrflow}—a probabilistic super-resolution model that learns conditional distributions from low- to high-resolution data using normalizing flows, enabling UQ in super-resolution tasks.
While these methods offer various forms of UQ, they lack the ability to produce interval predictions with guaranteed statistical coverage. To address this, Gopakumar~\cite{gopakumar2024uncertainty} et al. propose a general conformal prediction framework that enables uncertainty quantification for arbitrary surrogate models. Their approach provides statistically guaranteed prediction intervals. However, their work has only been demonstrated on 2D outputs, and, like the other methods, does not differentiate between distinct sources of uncertainty.


\section{Background: Uncertainty}
In this section, we introduce background on uncertainty by clarifying its definition, categorizing different types of uncertainty, and discussing their respective sources within the context of ensemble datasets. Understanding these aspects provides a foundation for effectively quantifying and visualizing uncertainty using our proposed  approach.

\subsection{Definition of Uncertainty}
Uncertainty in scientific visualization refers to "the \textbf{error, confidence, and variation} associated with the data," reflecting the degree of trust that one can place in the values of the data and the processes that visualize them~\cite{johnson2004top, potter2008towards}. 
Due to inherent limitations, no simulation can perfectly model real-world phenomena, nor can any surrogate model exactly reproduce the outputs of ensemble simulations. 
Consequently, there is always some degree of error arising from the generalizations made by these models. 
Thus, effective quantification and visualization of such uncertainty is necessary.

\subsection{Types of Uncertainty in Ensemble Datasets}
Uncertainty is broadly classified into two types: aleatoric uncertainty and epistemic uncertainty, both of which are essential to the understanding of prediction outcomes. Below we introduce each term. 

\subsubsection{Aleatoric uncertainty}
Aleatoric uncertainty (also known as stochastic or data uncertainty) refers to the inherent variability or randomness present within the data itself. 
It typically originates from inherent randomness during data acquisition.
Importantly, aleatoric uncertainty can only be mitigated by improving measurement precision or sample quality but cannot be reduced simply by collecting additional data or refining the model structure.
However, it can still be quantified explicitly using appropriate probabilistic or statistical methods.
For our proposed {\sysname}, aleatoric uncertainty primarily arises from the simulation-based ensemble data. Specifically, there are four main sources:
\begin{itemize}
\item \textbf{Stochastic Simulations}: some simulations are governed by stochastic equations. Thus, with identical input parameters, repeated runs yield different outputs due to intrinsic stochasticity. 
\item \textbf{Limited Resolution}: simulations discretize continuous physical phenomena onto finite grids or discrete time steps. If the spatial or temporal resolution is coarse, small-scale features or fluctuations cannot be accurately captured. Consequently, uncertainty manifests prominently in low-resolution outputs, particularly in regions with complex, rapidly varying features.
\item \textbf{Truncation and Rounding Errors}: numerical simulations inherently involve truncation and rounding errors during computation. 
\item \textbf{Simulation Systematic Error}: simulations often include simplifying assumptions or approximations that introduce systematic deviations. For our {\sysname}, such systematic deviations become intrinsic characteristics of the ensemble data, representing irreducible uncertainty from the modeling perspective.
\end{itemize} 
\subsubsection{Epistemic uncertainty}
Epistemic uncertainty (also called systematic or model uncertainty) arises from incomplete knowledge about the data or the limitations of the modeling approach itself. In other words, it is derived from what we do not currently know but could potentially learn through additional information. 
Unlike aleatoric uncertainty, epistemic uncertainty can be reduced by collecting more training data and improving model structure. 
In the context of our {\sysname}, epistemic uncertainty primarily originates from two sources:
\begin{itemize}
\item \textbf{Imperfections in surrogate model}: any deep learning model, including {\sysname}, inherently involves simplifying assumptions and approximation errors. Simulation outputs often have high dimensionality, requiring surrogate models to learn a large number of parameters. Consequently, these models are prone to overfitting, potentially leading to erroneous predictions.
\item \textbf{Insufficient Training Data}: ensemble datasets typically contain relatively few data points due to the high computational cost associated with running simulations. Limited data coverage results in epistemic uncertainty, as the surrogate model must extrapolate or interpolate in regions with sparse or no data.
\end{itemize} 

In our work, we leverage {\sysname} to explicitly distinguish between these two types of uncertainty. Clearly separating them informs domain experts whether uncertainty primarily arises from inherent data variability or modeling limitations, guiding more effective strategies to reduce uncertainty and enabling better-informed decisions.

\section{Method}
Existing surrogate methods do not yet provide effective uncertainty quantification, making it difficult for researchers to evaluate model reliability in high-dimensional predictions and complex parameter spaces. As a result, they often resort to extensive computations and experiments or simply ignore it, wasting considerable time and resources. 
To address this challenge, we propose {\sysname}, a neural network-based surrogate system that integrates uncertainty quantification capabilities with an interactive visualization interface. 
By focusing on regions of high uncertainty, {\sysname} enables researchers to pinpoint critical areas more effectively, thereby significantly improving the efficiency of the exploration of complex simulations.

\subsection{Overview}
\vspace{-8pt}
\begin{figure}[htp]
    \centering
    \includegraphics[width=\columnwidth]{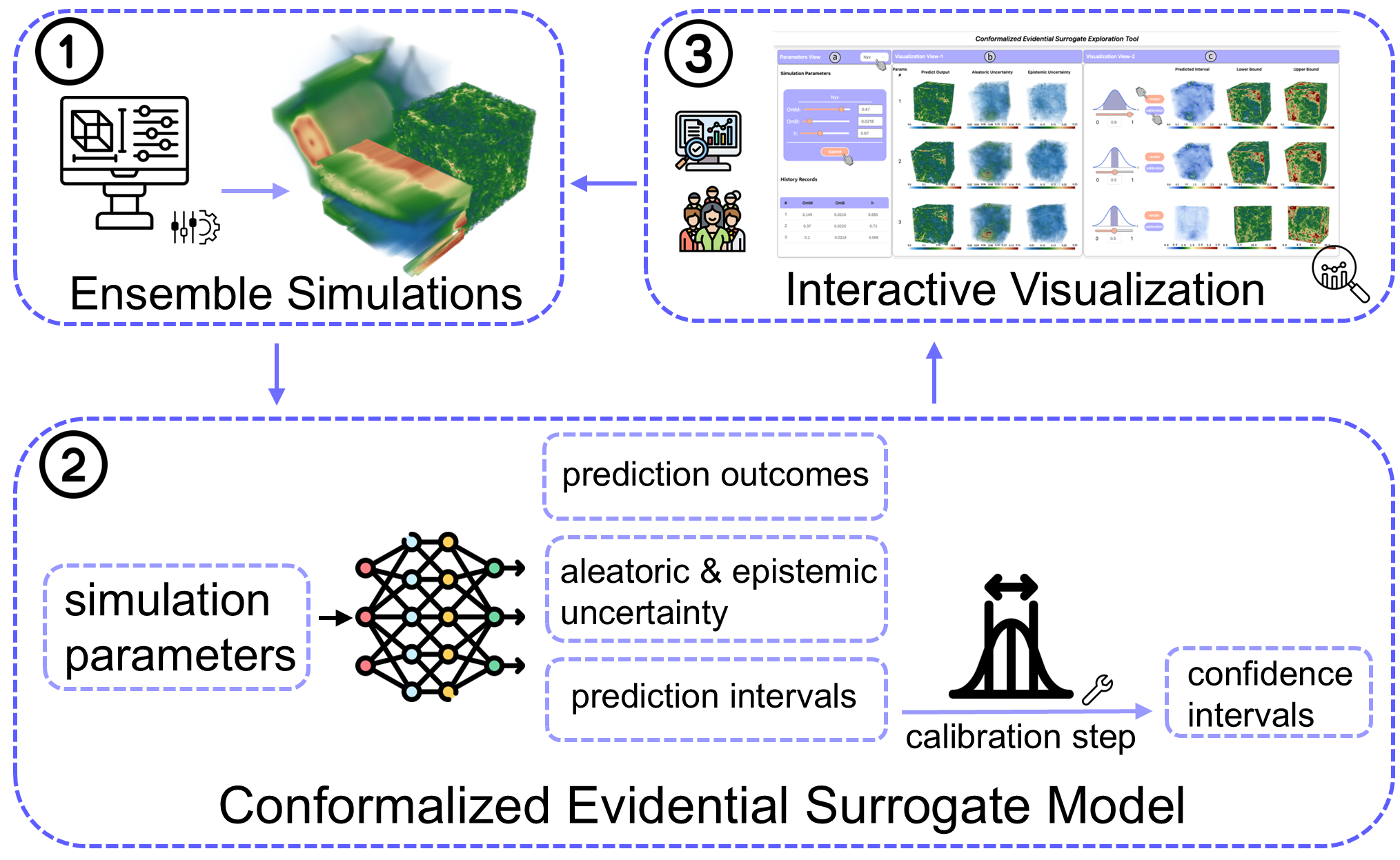}
    \vspace{-12pt}
    \caption{Overview of our approach. (1) Simulations are conducted with different simulation parameters, yielding a range of output data. (2) These data and their associated parameters are then used to train our Conformal Evidential Surrogate Model, which provides uncertainty quantification. (3) An interactive visualization interface allows users to explore parameters, predicted results, and associated uncertainties.}
    \label{fig:overview}
    \vspace{-8pt}
\end{figure}

\Cref{fig:overview} shows an overview of our framework. We propose {\sysname}, a conformalized evidential surrogate model for UQ, trained on simulation-based ensemble datasets. 
\textbf{First}, we train an evidential deep regression (EDR)-based surrogate on simulation parameters and their corresponding simulation outputs. 
Once this model is well-trained, scientists can predict unseen simulation results and their associated uncertainties of different types, with new simulation parameters as inputs. The model can also produce predictive intervals for simulation output. 
\textbf{Second}, to ensure theoretically guaranteed coverage of these intervals, we employ a conformal prediction-based algorithm to calibrate the initially generated intervals. The calibrated intervals not only satisfy the required coverage level but also attain shorter widths.
\textbf{Third}, we integrate our {\sysname} into an interactive visualization interface, allowing users to visualize and explore predicted outputs and associated uncertainties. 

\subsection{Deep Evidential Regression Surrogate} 
\label{sect:evidential surrogate}
Our proposed approach, {\sysname}, leverages the Deep Evidential Regression (DER) framework introduced by Amini et al.~\cite{amini2020deep}. DER is specifically designed for regression tasks, providing a principled way to simultaneously capture both aleatoric uncertainty and epistemic uncertainty. 
In this section, we discuss details about DER.

\subsubsection{Evidential Distribution Modeling of Uncertainty} \label{sect:evidential distribution}
Due to inherent uncertainties, the same set of simulation parameters $\mathbf{x} \in \mathbb{R}^{d}$ may yield multiple plausible outputs $\mathbf{y}_i \in \mathbb{R}^{D \times H \times W}$ (for $i = 1,2,\ldots$), 
or even a continuous range of possible values, where $D$, $H$, and $W$ denote the depth, height, and width of the simulation outputs, respectively. 
For simplicity of exposition, we illustrate our method using a one-dimensional scaler field output value $y$. 
The extension to high-dimensional outputs $\mathbf{y}$ is conceptually straightforward and follows the same reasoning. Consequently, simulation data can be naturally viewed as samples drawn from an underlying conditional distribution $p(y|\mathbf{x})$, where $\mathbf{x}$ denotes the input simulation parameters and $y$ represents the corresponding simulation outputs. 
A common modeling strategy assumes the target values $y$ are independently and identically distributed (i.i.d.) samples from a Gaussian distribution:
\vspace{-5pt}
\begin{equation}
\begin{gathered} 
y|\mathbf{x} \sim \mathcal{N}(\mu, \sigma^2), 
\end{gathered}
\label{eq:1}
\vspace{-5pt}
\end{equation}
where $\mu$ and $\sigma^2$ denote the mean and variance, respectively. While this Gaussian assumption effectively captures aleatoric uncertainty through the variance term, it fails to account for predictive epistemic uncertainty (the model’s uncertainty about its predictions). Alternative methods, such as dropout~\cite{gal2016dropout} or deep ensembles~\cite{lakshminarayanan2017simple}, can approximate epistemic uncertainty by producing variations in predictive outcomes, but these approaches slow down inference due to multiple runs.

To efficiently approximate epistemic uncertainty, DER introduces evidential priors over the parameters of the Gaussian distribution, thereby modeling both aleatoric and epistemic uncertainties without requiring sampling during inference.
Rather than directly predicting fixed Gaussian parameters $(\mu, \sigma^2)$, the evidential framework places a higher-order, Normal Inverse-Gamma (NIG) prior on $(\mu, \sigma^2)$. Formally,
\vspace{-4pt}
\begin{equation}
\begin{gathered} 
y|\mathbf{x} \sim \mathcal{N}(\mu, \sigma^2), \\
\mu \sim \mathcal{N}(\gamma, \sigma^2 \nu^{-1}), \quad \sigma^2 \sim \Gamma^{-1}(\alpha, \beta),
\end{gathered}
\label{eq:2}
\vspace{-5pt}
\end{equation}
where \(\Gamma^{-1}(\cdot)\) is the Inverse-gamma function, with \(\gamma \in \mathbb{R}, \nu > 0, \alpha > 1, \beta > 0\). 
Under this hierarchical formulation, the variance term $\sigma^2$
directly captures the inherent data uncertainty (aleatoric uncertainty). Meanwhile, modeling the mean parameter $\mu$ as a distribution rather than a fixed value naturally encodes the uncertainty arising from the model's limited knowledge or incomplete training data (epistemic uncertainty). Thus, the NIG prior effectively integrates both types of uncertainty within a single coherent probabilistic framework.

Equivalently, the above hierarchical formulation implies that the posterior distribution $q$ of the mean $\mu$ and variance $\sigma^2$ follows a Normal Inverse-Gamma (NIG) distribution:
\vspace{-3pt}
\begin{equation}
q(\mu, \sigma^2) = \text{NIG}(\mu, \sigma^2 \mid \gamma, \nu, \alpha, \beta).
\label{eq:3}
\vspace{-3pt}
\end{equation}
Then the joint probability density function of the NIG can be obtained:
\vspace{-3pt}
\begin{equation}
\scalebox{0.822}{$
\begin{aligned} 
    p(\underbrace{\mu, \sigma^2}_{\mathbf{\theta}} \mid \underbrace{\gamma, \nu, \alpha, \beta}_{\mathbf{m}} \big) 
    &= \frac{\beta^\alpha \sqrt{\nu}}{\Gamma(\alpha) \sqrt{2\pi \sigma^2}} 
    \left( \frac{1}{\sigma^2} \right)^{\alpha+1} 
    \exp\left( -\frac{2\beta + \nu (\gamma - \mu)^2}{2\sigma^2} \right),
\end{aligned}
$}
\label{eq:joint pdf of NIG}
\vspace{-5pt}
\end{equation}
where \(\Gamma(\cdot)\) is the gamma function, \(\mathbf{\theta} = (\mu, \sigma^2)\), and \(\mathbf{m} = (\gamma,  \nu, \alpha, \beta)\). 
In this formulation, the NIG can act as a higher-order distribution placed over the parameters $(\mu, \sigma^2)$ of the Gaussian distribution, which serves as the lower-order likelihood distribution that the observed data is drawn from. Drawing multiple samples $\mathbf{\theta}_j = (\mu_j, \sigma_j^2)$ (for $j = 1,2,\ldots$) from the NIG distribution corresponds to generating a family of possible Gaussian distributions, each capable of producing the observed data. Thus, the NIG hyperparameters $(\gamma, \nu, \alpha, \beta)$ encode not only the aleatoric uncertainty originally captured by the Gaussian parameters $(\mu, \sigma^2)$, but also the epistemic uncertainty arising from variability in predictive output $\mu$.

Given the posterior distribution in~\cref{eq:joint pdf of NIG} and the Gaussian likelihood, we can further derive an explicit form of the predictive distribution $p(y_i\mid \mathbf{m})$. This distribution serves as the foundation later for constructing interval predictions for the simulation output $y_i$. Importantly, this derivation highlights the mathematical rigor of evidential formulation, naturally bridging the NIG prior and the robust Student-t distribution (see detailed derivation in supplemental material):
\vspace{-2pt}
\begin{equation}
\begin{aligned}
p(y_i\mid \mathbf{m}) &= \int^{\infty}_{\sigma^2=0} \int^{\infty}_{\mu=-\infty} 
p(y_i \mid \mu, \sigma^2) p(\mu, \sigma^2 \mid \mathbf{m}) \,d\mu \,d\sigma^2 \\
&= \text{St} \left( y_i; \gamma, \frac{\beta(1+\nu)}{\nu\alpha}, 2\alpha \right) \\
\end{aligned}
\label{eq:5}
\vspace{-1pt}
\end{equation}
where \(
\text{St}\left(y;\,\mu_{St}, \sigma_{St}^2, \nu_{St}\right)
\) is the Student-t distribution, $\mu_{St}$ is the location parameter, $\sigma_{St}^2$ is the scale parameter, and $\nu_{St}$ represents the degrees of freedom. Compared to the Gaussian distribution, Student-t distribution has heavier tails, making it more robust to outliers.

\subsubsection{Prediction and Uncertainty Estimation}
\label{sect:evidential predictions}
The aleatoric uncertainty is representative of
unknowns that differ each time we run the same experiment. The epistemic uncertainty describes the estimated uncertainty in the prediction. 

Given the Normal-Inverse-Gamma (NIG) prior distribution and the Student-t predictive distribution, the four hyperparameters \(\mathbf{m} = (\gamma, \nu, \alpha, \beta)\) can be used to derive output prediction, aleatoric uncertainty, epistemic uncertainty, and initial predictive intervals as follows (see detailed derivation in supplemental material):
\begin{itemize}
    \item Simulation output prediction: \( \mathbb{E}[\mu] = \gamma \)
    \item Aleatoric uncertainty: \( \mathbb{E}[\sigma^2] = \frac{\beta}{\alpha - 1} \)
    \item Epistemic uncertainty: \( \text{Var}[\mu] = \frac{\beta}{\nu(\alpha - 1)} \)
    \item A prediction interval for \( y_i \) at a confidence level \( 1 - \delta \) is given by:
    \[
\left[ \gamma - t_{2\alpha,1-\frac{\delta}{2}} \sqrt{\frac{\beta (1+\nu)}{\nu \alpha}}, \gamma + t_{2\alpha,1-\frac{\delta}{2}} \sqrt{\frac{\beta (1+\nu)}{\nu \alpha}} \right]
\]
\end{itemize}

\subsubsection{Model Architecture} \label{sect:Model Architecture}

\begin{figure}[htp]
    \centering
    \includegraphics[width=\columnwidth]{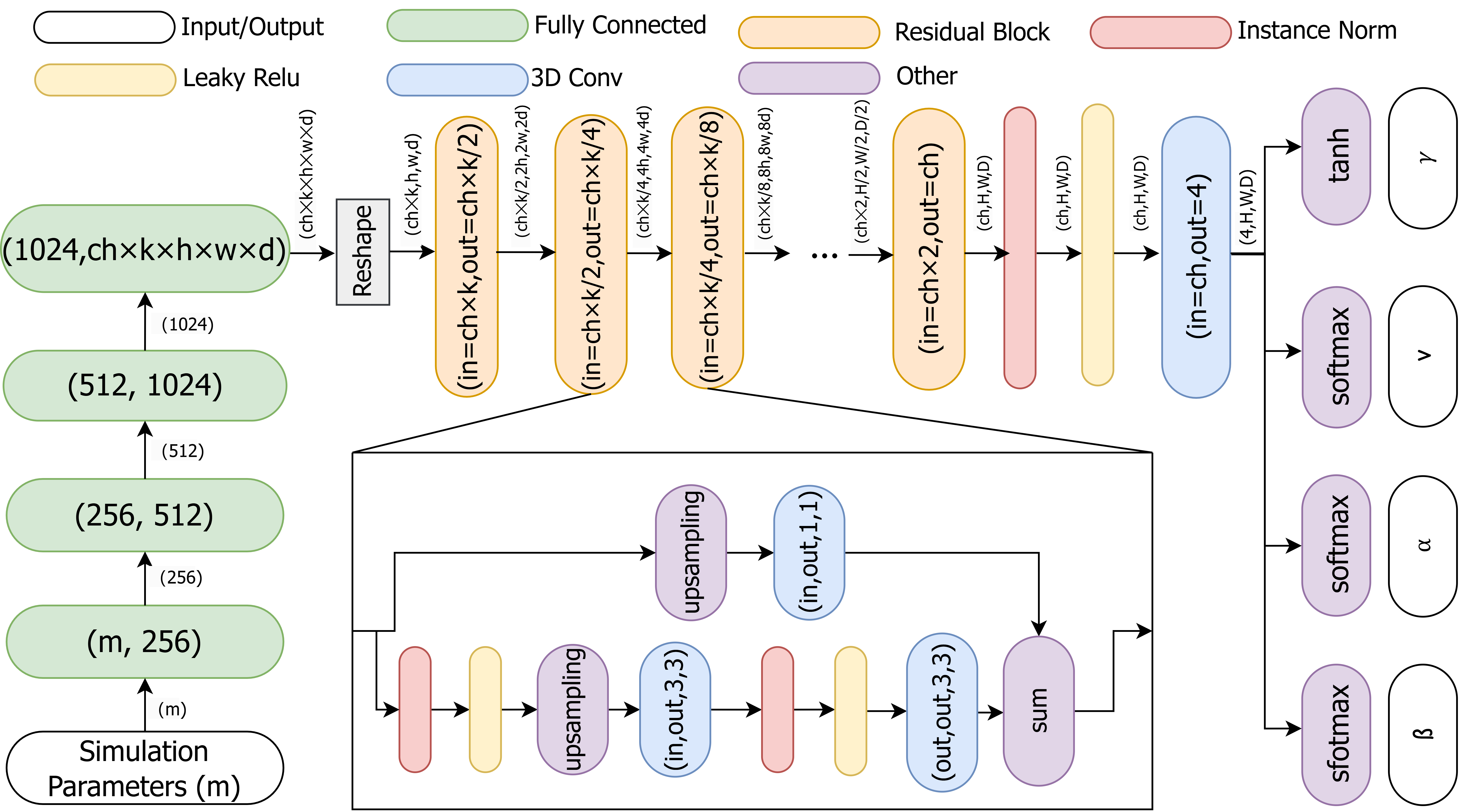}
    \vspace{-12pt}
    \caption{{Architecture of {\sysname}}, which generates the hyperparameters \((\gamma,\nu,\alpha,\beta)\) of the higher-order evidential distribution given input simulation parameters. The size of our model is defined by \(\text{ch}\) and \(k\). Here, \(\text{ch}\) controls the number of channels in the intermediate layers, while \(k\) determines how many times we upsample the low-resolution tensor \((h,w,d)\) to match the final output resolution \((H,W,D)\).}
    \label{fig:network}
    \vspace{-6pt}
\end{figure}

To model the evidential distribution of simulation outputs, our {\sysname}  utilizes a CNN-based architecture to predict hyperparameter \(\mathbf{m}=(\gamma, \nu, \alpha, \beta)\).
As illustrated in~\cref{fig:network}, the architecture first encodes the simulation input parameters \(\mathbf{x}\) through fully connected layers, producing a latent representation. This latent vector is then reshaped into a low-resolution 3D tensor suitable for convolutional processing. Subsequently, several residual blocks \cite{he2016deep} perform successive 3D convolutions and upsampling steps, gradually producing a high-resolution feature map.

In the final layer, we generate a four-channel feature map corresponding to the four hyperparameters, applying specific activation functions to satisfy parameter constraints: a \texttt{tanh} activation for \(\gamma\) ensuring it lies within \([-1, 1]\), and a \texttt{softplus} activation for \((\nu,\alpha,\beta)\) ensuring positivity, with an additional offset of \(+1\) for \(\alpha\) to ensure \(\alpha > 1\).

\subsubsection{Loss function}
To train our model introduced in~\cref{sect:Model Architecture} under the evidential framework, we use a combination loss function that ensures the model can learn the evidential distribution. Given simulation parameters \(\mathbf{x}\) and corresponding outputs \(y\), our loss consists of three components: a negative log-likelihood loss ($\mathcal{L}^{\text{NLL}}$) that maximizes model fit to the data, and an evidence regularization loss (
$\mathcal{L}^{\text{R}}$) to penalize overconfident or misleading evidence~\cite{amini2020deep}. Additionally, we incorporate an extra regularization term $\mathcal{L}^{\text{U}}$ to further enhance robustness and stability~\cite{wu2024evidence}.

First, the negative log-likelihood (NLL) loss is commonly used in probabilistic models to optimize parameters by maximizing the likelihood of observed data under the predicted distribution. Based on the marginal likelihood of the output described in \cref{eq:5}, the NLL loss can be explicitly formulated as:
\vspace{-3pt}
\begin{equation}
\begin{aligned}
\mathcal{L}_i^\text{NLL} &= -log p(y_i \mid \mathbf{m})\\
&= \left(\alpha + \frac{1}{2}\right)\log\left((y_i - \gamma)^2\nu + \Omega\right) \\
\vspace{-1pt}
&+\frac{1}{2}\log(\frac{\pi}{\nu}) - \alpha\log(\Omega) 
+ \log\left(\frac{\Gamma(\alpha)}{\Gamma(\alpha + \frac{1}{2})}\right).
\end{aligned}
\label{eq:6}
\vspace{-3pt}
\end{equation}
where \(\Omega = 2\beta(1+\nu)\), and \(\alpha, \beta, \nu\) are parameters of the NIG distribution. This loss ensures that the predicted evidential distribution aligns closely with the observed data.

Second, we incorporate an evidence regularization term to better capture epistemic uncertainty. 
Originally proposed by Amini et al.~\cite{amini2020deep}, this term penalizes 
overly confident predictions with large errors and can be defined as follows:
\vspace{-3pt}
\begin{equation}
\mathcal{L}_i^R = |y_i - \gamma| \cdot (2\nu + \alpha).
\label{eq:7}
\vspace{-3pt}
\end{equation}
Although the original work introduced this formulation, it did not provide a detailed explanation. 
We include a more comprehensive discussion of this regularization term 
in the supplemental material.

Third, we incorporate the non-saturating uncertainty regularization term proposed by Wu et al.~\cite{wu2024evidence}
to address the evidence contraction issue. Evidence contraction occurs when parameters such as \(\nu\) approach zero,
leading the model to underfit the data and thereby produce unreliable uncertainty estimates.
The non-saturating uncertainty regularization loss is defined as:
\vspace{-6pt}
\begin{equation}
\mathcal{L}_i^U = (y_i - \gamma)^2 \frac{\nu (\alpha - 1)}{\beta (\nu + 1)},
\label{eq:8}
\vspace{-6pt}
\end{equation}
where \(\frac{\nu (\alpha - 1)}{\beta (\nu + 1)}\) represents the inverse of the sum of two uncertainties.

In summary, our total loss \(\mathcal{L}_i\) integrates three key components to achieve effective evidential regression: 
\vspace{-3pt}
\begin{equation}
\begin{gathered} 
\mathcal{L}_i = \mathcal{L}_i^\text{NLL} + \lambda \mathcal{L}_i^R + \xi \mathcal{L}_i^U,
\end{gathered}
\label{eq:9}
\vspace{-2pt}
\end{equation}
where \(\lambda\) and \(\xi\) control the trade-offs among model fit, evidence regularization, and uncertainty calibration, ensuring that the model achieves both accuracy and reliable uncertainty quantification. 
%

\subsection{Conformal prediction}
As discussed in~\cref{sect:evidential surrogate}, the trained evidential model directly provides predictive intervals derived from the evidential distribution. However, these intervals do not inherently possess finite-sample coverage guarantees—meaning they may not contain the true outcomes with a specified probability. To rigorously address this limitation, we introduce a calibration step based on conformal prediction, which leverages additional calibration data to refine the initial intervals. 
This step ensures finite-sample coverage guarantees at any predefined coverage level.

\subsubsection{Definition of Coverage Guarantee}
Given a coverage level $\delta \in (0,1)$ and data pairs $(X,Y)\sim \mathcal{P}$ drawn from an unknown distribution $\mathcal{P}$, a prediction interval $\mathcal{C}_{1-\delta}(X)$ satisfies the \emph{coverage guarantee} if:
\vspace{-3pt}
\begin{equation}
    \mathbb{P}\left(Y \in \mathcal{C}_{1-\delta}(X)\right) \geq 1 - \delta.
    \label{eq:coverage}
\vspace{-3pt}
\end{equation}
In other words, the constructed prediction intervals will cover the true outcomes with probability at least $1-\delta$ over repeated sampling from the underlying distribution. Practically, there exists a natural trade-off between the width (or volume) of prediction intervals and their coverage probability: intervals can always be widened to trivially achieve high coverage, but narrower intervals are usually preferable. Conformal prediction provides a principled approach to optimize interval width while simultaneously maintaining rigorous coverage guarantees. Below we describe how this can be done. 

\subsubsection{Dataset Splitting}
Applying conformal prediction requires splitting our dataset into two independent subsets to ensure the validity of coverage guarantees:
\begin{itemize}
    \item A \textbf{training set} $\mathcal{I}_1$, used to fit the evidential model. This set determines the model parameters and initial predictive intervals for simulation outputs.
    \item A \textbf{calibration set} $\mathcal{I}_2$, used to calculate non-conformity scores, which quantify the discrepancy between the evidential model predictions and ground-truth ensemble data. These scores are subsequently used to calibrate and refine the prediction intervals.
\end{itemize}
This strict separation of roles between $\mathcal{I}_1$ and $\mathcal{I}_2$ is crucial, as the independence of calibration data is essential to guarantee the finite-sample validity of conformal intervals.

Moreover, conformal prediction relies on exchangeability (see the formal definition in supplemental material), which is weaker than the common i.i.d. assumption. Our {\sysname} dataset—pairs of input parameters and corresponding simulation volumes—was generated under i.i.d.\ conditions, thus inherently satisfying exchangeability. Consequently, our dataset meets the requirements for conformal prediction.

\subsubsection{Non-Conformity Score Design}
In this work, we treat each voxel in the 3D volumetric data as an individual one-dimensional regression target and apply conformal prediction to each voxel independently. Specifically, for each voxel $v$, we construct a calibration dataset ${(X_i, Y_i^v)}_{i=1}^n$, where $Y_i^v$ is the ground-truth simulation value at voxel $v$. Subsequently, we use a split conformal procedure to calibrate the initial predictive interval and obtain the refined interval $\hat{C}^v(X)$ at a confidence level of $1-\alpha$. For notational simplicity, we use the scalar representation $Y_i$ throughout the remainder of this section, implicitly referring to an individual voxel's one-dimensional output.

As described in~\cref{sect:evidential predictions}, the evidential framework provides an initial predictive interval of simulation output $Y_i$, denoted as $[\hat{q}_{\alpha_{\text{lo}}}(X_i), \hat{q}_{\alpha_{\text{hi}}}(X_i)]$, based on the inferred Student-t distribution. 

To calibrate this interval, we define two separate non-conformity scores for each data point $(X_i, Y_i)$ in the calibration set $\mathcal{I}_2$:
\vspace{-3pt}
\begin{equation}
    E_i^{\text{lo}} = \hat{q}_{\alpha_{\text{lo}}}(X_i) - Y_i,\quad 
    E_i^{\text{hi}} = Y_i - \hat{q}_{\alpha_{\text{hi}}}(X_i).
\label{eq:nonconformity_scores}
\vspace{-7pt}
\end{equation}
Here, $E_i^{\text{lo}}$ measures the deviation of the observed value from the lower bound, while $E_i^{\text{hi}}$ measures the deviation from the upper bound. Positive values of $E_i^{\text{lo}}$ or $E_i^{\text{hi}}$ indicate violations of the initial predictive interval, meaning the corresponding bound underestimated the uncertainty, resulting in ground-truth values falling outside the interval. Conversely, negative values imply a successful coverage by the initial interval. Predominantly negative values across the calibration set could indicate that intervals are excessively wide, potentially reflecting overly cautious and conservative predictions.

\subsubsection{Conformal Calibration Procedure}
We perform the following calibration steps to construct final prediction intervals with rigorous coverage guarantees: 

\begin{algorithm}
\caption{Conformal Calibration Procedure}\label{alg:conformal_calibration}
\begin{algorithmic}[1]
\State \textbf{Input:} Training set $\mathcal{I}_1$, calibration set $\mathcal{I}_2$, significance level $\alpha$, new test point $X_{n+1}$
\State \textbf{Output:} Calibrated prediction interval $\mathcal{C}_{1-\alpha}(X_{n+1})$
\Procedure{ConformalCalibration}{$\mathcal{I}_1$, $\mathcal{I}_2$, $\alpha$, $X_{n+1}$}
    \State Fit evidential model on $\mathcal{I}_1$ to get $[\hat{q}_{\alpha_{\text{lo}}}(X), \hat{q}_{\alpha_{\text{hi}}}(X)]$
    \For{each $(X_i, Y_i) \in \mathcal{I}_2$}
        \State $E_i^{\text{lo}} \gets \hat{q}_{\alpha_{\text{lo}}}(X_i)-Y_i$
        \State $E_i^{\text{hi}} \gets Y_i - \hat{q}_{\alpha_{\text{hi}}}(X_i)$
    \EndFor
    \State $Q_{1-\alpha}^{s} \gets \text{Quantile}\left({E_i^{s}}, (1-\alpha)(1+\frac{1}{|\mathcal{I}2|})\right),\quad s\in{\{\text{lo},\text{hi}\}}$
    \State $\mathcal{C}_{1-\alpha}(X_{n+1}) \gets \left[\hat{q}_{\alpha_{\text{lo}}}(X_{n+1}) - Q_{1-\alpha}^{\text{lo}}, \hat{q}_{\alpha_{\text{hi}}}(X_{n+1}) + Q_{1-\alpha}^{\text{hi}}\right]$
    \State \textbf{return} $\mathcal{C}_{1-\alpha}(X_{n+1})$
\EndProcedure
\end{algorithmic}
\end{algorithm}

Firstly, given the calibration dataset $\mathcal{I}_2$, we calculate the non-conformity scores defined in~\cref{eq:nonconformity_scores}, quantifying the discrepancy between the ground-truth values and the evidential prediction intervals.

Secondly, using these scores, we compute calibration thresholds separately for the lower and upper bounds. Specifically, we obtain $(1-\alpha)(1 + 1/|\mathcal{I}_2|)$-th empirical quantiles from the calibration set: 
\vspace{-4pt}
\begin{equation}
\begin{gathered} 
Q_{1-\alpha}^{s}(E^{s}, \mathcal{I}_2) = \text{Quantile}\left({E_i^{s} : i \in \mathcal{I}_2}, (1-\alpha)(1 + 1/|\mathcal{I}_2|)\right), \\
s\in{\{\text{lo},\text{hi}\}}. 
\end{gathered}
\label{eq:quantile_thresholds}
\vspace{-3pt}
\end{equation}
These empirical quantiles are then used to calibrate and construct the final prediction intervals, ensuring strict coverage guarantees.

Thirdly, with the empirical quantiles obtained from the calibration set~\cref{eq:quantile_thresholds}, we construct the final calibrated prediction interval for a new input $X_{n+1}$ as follows:
\vspace{-5pt}
\begin{equation}
\mathcal{C}_{1-\alpha}(X_{n+1}) = \left[
\hat{q}_{\alpha_{\text{lo}}}(X_{n+1}) - Q_{1-\alpha}^{\text{lo}},\;
\hat{q}_{\alpha_{\text{hi}}}(X_{n+1}) + Q_{1-\alpha}^{\text{hi}}
\right].
\label{eq:calibrated_interval}
\vspace{-3pt}
\end{equation}

This calibration procedure refines the initial evidential intervals, ensuring finite-sample coverage guarantees. It prevents intervals from becoming overly conservative or excessively narrow, achieving a principled balance between coverage probability and interval width. From conformal prediction theory~\cite{vovk2005algorithmic, romano2019conformalized, angelopoulos2021gentle}, we can show that the resulting
intervals satisfy the marginal coverage guarantee in~\cref{eq:coverage}. The following theorem
further establishes this result:

\begin{theorem}
\label{thm:coverage}
If $(X_i, Y_i)$, $i = 1, \ldots, n+1$, are exchangeable, then the prediction interval \(\mathcal{C}(X_{n+1})\) constructed by our {\sysname} satisfies
\[
  \Pr\bigl\{ Y_{n+1} \in C(X_{n+1}) \bigr\} \;\ge\; 1 \;-\; \alpha.
\]
\end{theorem}
\vspace{-3pt}
Moreover, if the conformity scores $E_i$ are almost surely distinct, then the prediction interval is nearly perfectly calibrated. In particular, we have:
\vspace{-2pt}
\[
\Pr\bigl\{Y_{n+1} \in C(X_{n+1})\bigr\}
\;\le\;
1 \;-\; \alpha
\;+\;
\frac{1}{\lvert \mathcal{I}_2\rvert + 1}\,.
\]
\vspace{-8pt}

For a complete theoretical treatment, see supplemental material. 
Here, we focus on providing an intuitive understanding of the calibration procedure.
The key idea behind conformal prediction is that the calibration dataset and future unseen data are independently drawn from the same distribution. 
As a result, the distribution of prediction errors—measured via non-conformity scores—on the calibration set serves as a good approximation for the errors on unseen data.
This enables us to use empirical quantiles from the calibration set as thresholds for future predictions.
In other words, the quantile thresholds derived from the calibration dataset can be viewed as representative thresholds for future predictions. However, since the calibration dataset is finite, it inevitably differs from the infinite future test dataset. 
To correct for these finite-sample effects, the conformal method incorporates minor adjustments—such as using the
\((1-\alpha)(1 + 1/|\mathcal{I}_2|)\)-th when computing quantiles—to ensure mathematically valid finite-sample coverage guarantees.




\subsection{Visual Interface of ConfEviSurrogate Model}

We integrate our {\sysname} as the backend and develop a user-friendly interactive visual interface to facilitate the exploration of scientific phenomena. This intuitive, code-free interface allows scientists to focus on investigating predictions and uncertainties, rather than dealing with command-line codes.

\begin{figure}[htp]
    \centering    \includegraphics[width=1\columnwidth]{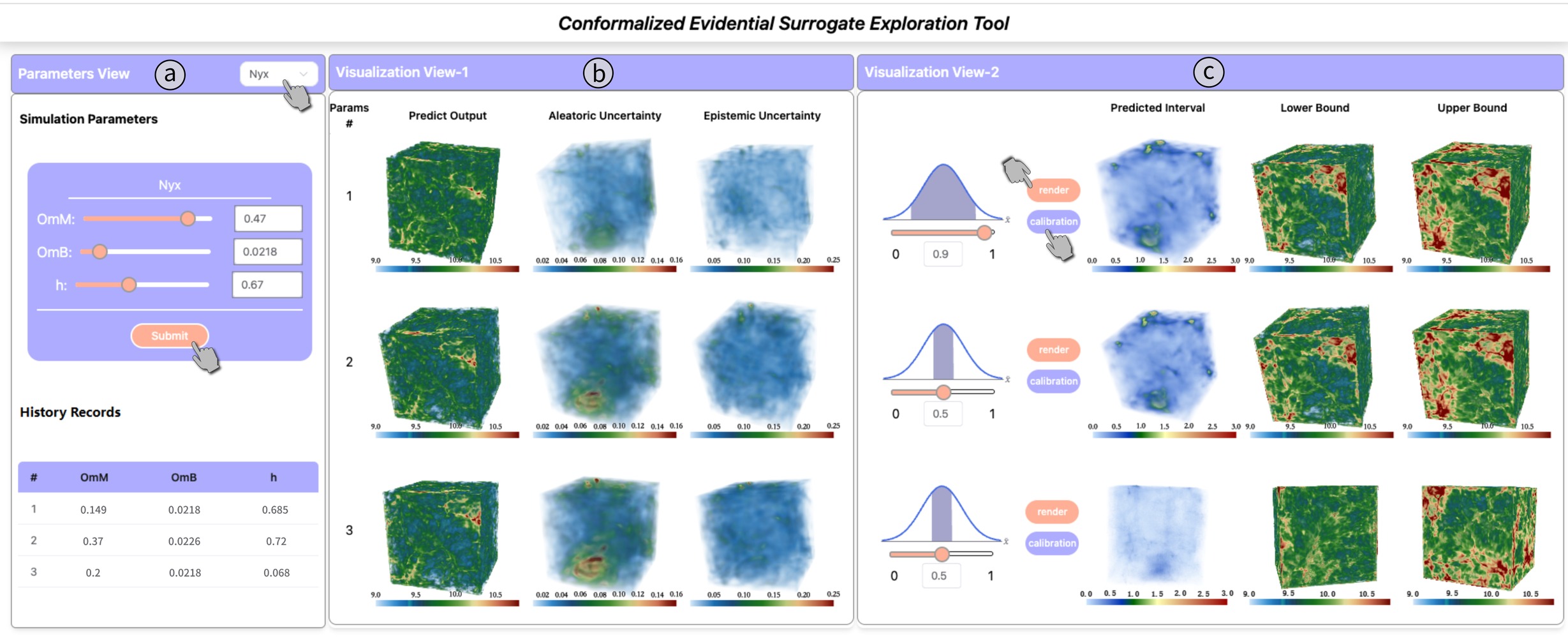}
    \caption{Visual interface for {\sysname} exploration. (a) Parameter View for selecting simulation parameters of interest. (b) Visualization View-1 for visualizing the predicted output and associated two uncertainties. (c) Visualization View-2 for visualizing the prediction intervals.}
    \label{fig:interface}
    \vspace{-18pt}
\end{figure}

Our visual interface facilitates the parameter exploration process. First, users can easily configure input parameters of interest through the Parameter View (\cref{fig:interface}a). Second, based on the selected inputs, we display our model’s predicted outputs along with the two associated types of uncertainty in Visualization View-1 (\cref{fig:interface}b). This not only allows users to clearly identify regions of high data noise (aleatoric uncertainty) and low model confidence (epistemic uncertainty), but also enables easy comparison across different ensemble data cases. To further support robust and trustworthy predictions, we present the corresponding prediction intervals in Visualization View-2 (\cref{fig:interface}c), including the lower and upper bounds of the predictive interval, as well as its width.
We also provide an interactive calibration slider for the post-processing calibration step, which can further tighten the interval bands. These recalibrated intervals are theoretically guaranteed to contain the true outcomes with the user-specified confidence level. Together, these views form a cohesive, interactive interface for exploring the relationships between inputs, predictions, and uncertainties in a code-free environment.


\section{Results} \label{sect:Results}
We evaluate {\sysname}'s capabilities as a surrogate model for data generation, uncertainty quantification, and interval prediction.

\subsection{Dataset and Implementation}
Our proposed {\sysname} is evaluated using three scientific ensemble simulation datasets, as summarized in~\cref{tab:table-1}.

\textbf{Nyx}~\cite{almgren2013nyx} is a cosmological hydrodynamics simulation developed by Lawrence Berkeley National Laboratory. Following expert recommendations, we select three cosmological parameters as inputs: the Omega Matter parameter (OmM), the Omega Baryon parameter (OmB), and the Hubble parameter (h). The model predicts the log density field on a $256 \times 256 \times 256$ grid. We use 128 simulations for training, 200 for calibration, and 100 for testing.

\textbf{MPAS-Ocean}~\cite{ringler2013mpas} is a collection of simulation results generated by the MPAS-Ocean model from Los Alamos National Laboratory. We focus on four input parameters: Bulk wind stress Amplification ($BwsA$), Gent-McWilliams eddy transport coefficient ($GM$), Critical Bulk Richardson Number ($CbrN$), and Horizontal Viscosity ($HV$). The model output is defined on a $192 \times 96 \times 96$ grid. Due to the high simulation cost, only 90 samples are available: 70 for training and 20 for testing. Given the data scarcity, we evaluate only the evidential method on this dataset, without applying conformal prediction.

%

\textbf{CloverLeaf3D}~\cite{mallinson2013cloverleaf} is a proxy application developed by the Atomic Weapons Establishment to solve the 3D compressible Euler equations on a structured grid. Based on expert recommendations, we focus on two simulation parameters: density and energy, which define three distinct physical states under specific physical constraints. Each simulation produces a 3D energy scalar field with a spatial resolution of $64 \times 64 \times 64$. 
In total, we generate 2000 members for training, 2000 for calibration, and 400 for testing.

\begin{table}[tb]
\small
\caption{Simulation dataset name and simulation parameter range.}
\label{tab:table-1}
\centering
\begin{tabular}{c|c}
Dataset              & Simulation Parameter Range  \\ 
\hline
MPAS-Ocean & \makecell{$BwsA \in [0.0, 5.0]$, $GM \in [300.0, 1500.0]$,\\ $CbrN \in [0.25, 1.00]$, $HV \in [100.0, 300.0]$} \\
\hline
Nyx & \makecell{$OmM \in [0.12, 0.155]$, $OmB \in [0.0215, 0.0235]$, $h \in [0.55, 0.85]$} \\
\hline
\multirow{6}{*}{CloverLeaf3D} & State 1: Density 1 $\in [0.01, 1.0]$, Energy 1 $\in [0.75, 2.0]$ \\
& State 2: Density 2 $\in [0.5, 2.0]$, Energy 2 $\in [1.5, 3.5]$ \\
& State 3: Density 3 $\in [1.5, 3.0]$, Energy 3 $\in [4.0, 7.0]$ \\
& \multicolumn{1}{r}{\makecell{\textit{(Energy 1 $<$ Energy 2 $<$ Energy 3,} \\ \textit{Density 1 $<$ Density 2 $<$ Density 3, Density $<$ Energy)}}} \\
\end{tabular}%
\end{table}

Our {\sysname} is implemented in PyTorch$\footnote{https://pytorch.org}$ and trained using a single NVIDIA A100 GPU. The visual system is implemented with Vue.js$\footnote{https://vuejs.org/}$ for the frontend and Flask$\footnote{https://flask.palletsprojects.com/}$ for the backend server. VTK.js$\footnote{https://kitware.github.io/vtk-js/}$ is used for volume rendering of data. 

\subsection{Surrogate Prediction}
In this section, we evaluate {\sysname}'s performance in generating high-quality surrogate data given simulation parameters, using both voxel-level and image-level metrics.
For voxel-level assessment, we compute the Peak Signal-to-Noise Ratio (PSNR) between the surrogate-generated data and ground truth simulations. 
For image-level evaluation, we use the Structural Similarity Index Measure (SSIM) between volume-rendered images derived from generated and actual simulation outputs. 
Higher PSNR and SSIM values indicate more accurate and visually faithful reconstructions.

We compare {\sysname} with two state-of-the-art surrogate models:
\begin{itemize}
  \item  VDL-Surrogate (shorted for VDL)~\cite{shi2022vdl}: a view-dependent surrogate model that leverages latent representations from selected viewpoints to enable efficient inference.
  \item SurroFlow~\cite{shen2024surroflow}: a flow-based surrogate model enabling invertible prediction and uncertainty quantification. 
\end{itemize}

\begin{table}[tb]
\small
\centering
\caption{Data dimension, model size, training time, and test time for each dataset.}
\label{tab:table-2}
\begin{tabular}{c|c|c|c|c|c}
{Dataset} & {\makecell{Model \\Name}} & {\makecell{Data \\Dimension}} & {$\downarrow$\makecell{Model \\Size}} & {$\downarrow$\makecell{Training \\Time}} & {$\downarrow$\makecell{Test \\Time}}\\
 \hline
 \multirow{3}*{\makecell{MPAS-\\Ocean}} & VDL & [192, 96, 96] & 0.63 GB & 139.8h & 21.0s \\
 ~ & {SurroFlow} & [192, 96, 12] & \textbf{47.93MB}  & \textbf{11.5h} & 0.81s \\
 ~ & {Ours} & [192, 96, 96] & 193.25MB  & 24h & \textbf{0.10s} \\
\hline
\multirow{3}*{Nyx} & VDL & [256, 256, 256] & 1.98 GB & 82.7h & 9.20s \\
 ~ & {SurroFlow} & [128, 128, 128] & \textbf{42.93MB} & 47.5h & 0.90s \\
 ~ & {Ours} & [256, 256, 256] & 227.03 MB & \textbf{24h} & \textbf{0.36s} \\
\hline
\multirow{3}*{\makecell{Clover-\\leaf3D}} & VDL & [64, 64, 64] &  0.68GB & 84h & 0.16s \\
 ~ & {SurroFlow} & [64, 64, 64] & 541.26MB  & 48h &  0.15s \\
 ~ & {Ours} & [64, 64, 64] & \textbf{212.19MB}  & \textbf{18h} & \textbf{0.03s} \\
\end{tabular}
\vspace{-12pt}
\end{table}

\begin{table}[t]
\small
\centering
\caption{PSNR and SSIM for {\sysname} and baselines' results.}
\label{tab:table-3}
\begin{tabular}{c|c|c|c|c}
Dataset & Method & Data Dimension & $\uparrow$PSNR & $\uparrow$SSIM  \\
\hline
\multirow{3}*{MPAS-Ocean} & VDL & [192, 96, 96] & 41.3098 & 0.9643 \\
  ~ & SurroFlow  & [192, 96, 12] & 46.6852 & 0.9950 \\
  ~ & Ours & [192, 96, 96] & \textbf{49.7259} & \textbf{0.9961} \\
\hline
\multirow{3}*{Nyx} & VDL & [256, 256, 256] & 35.7962 & 0.9180 \\
  ~ & SurroFlow  & [128, 128, 128] & 30.9263  & 0.8304 \\
  ~ & Ours & [256, 256, 256] & \textbf{37.0429} & \textbf{0.9312} \\
\hline
\multirow{3}*{Cloverleaf} & VDL & [64, 64, 64] &  38.3946 & 0.9326 \\
  ~ & SurroFlow  & [64, 64, 64] & 32.7111 & 0.8798 \\
  ~ & Ours & [64, 64, 64] & \textbf{47.7264} & \textbf{0.9817} \\
\end{tabular}
\vspace{-5pt}
\end{table}

\Cref{tab:table-2} summarizes the training data dimensions, model sizes, training time, and inference time across datasets. 
Notably, SurroFlow's normalizing flow architecture requires consistent data dimensionality across network layers, which leads to high memory usage. 
As a result, its model architecture is constrained by memory limitations, hindering its ability to generate high-resolution outputs. 
In contrast, both our {\sysname} and VDL successfully support high-resolution predictions.
Among the three models, {\sysname} achieves the smallest number of parameters, the shortest training time, and the fastest inference when operating on datasets of the same resolution. 
Although SurroFlow shows  shorter training time and smaller model sizes on certain datasets, this stems from operating on lower-dimensional data, which simplifies the prediction task and limits fair comparison.
\Cref{tab:table-3} further demonstrates that we achieve the highest PSNR and SSIM scores across all datasets. 
Importantly, our model outperforms both VDL and SurroFlow even though SurroFlow benefits from a simpler task due to its lower data dimensionality.
This superior performance can be attributed to two key design choices. 
First, our model leverages the negative log-likelihood (NLL) loss, whereas VDL uses a mean squared error (MSE)-based loss. 
As formally shown in the supplemental material, the NLL loss implicitly incorporates the MSE term, enabling more effective joint learning of accuracy and uncertainty. 
Second, although SurroFlow also adopts the NLL loss, its performance is limited by the architectural constraint of invertibility, which enforces a fixed network structure and restricts its capacity to capture complex data patterns.




\begin{figure}
    \centering
    \includegraphics[width=\columnwidth]{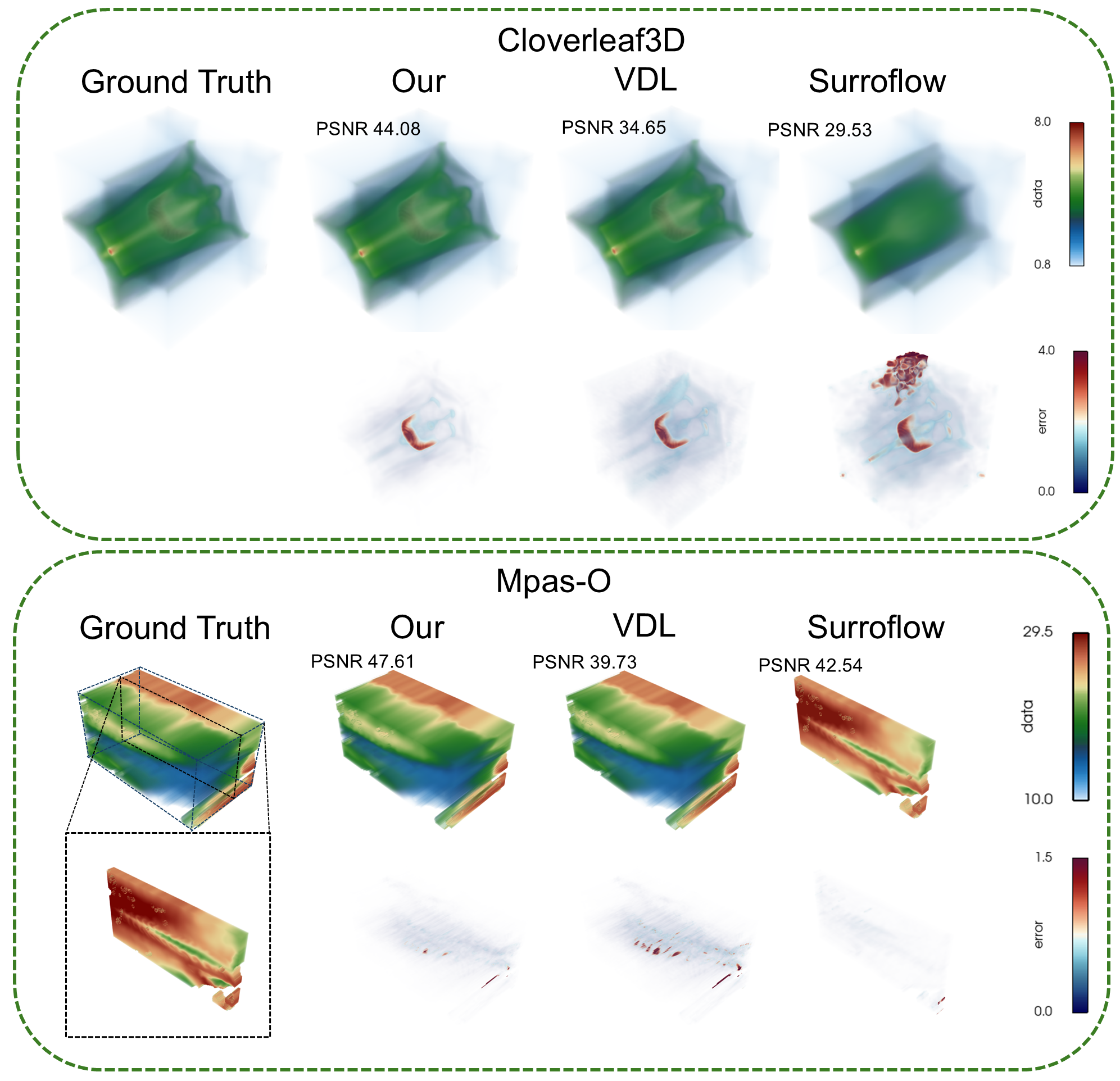}
    \caption{Volume rendering of Cloverleaf3D and Mpas-O data for predictive outputs (top row) and corresponding error maps (bottom row). In Mpas-O,  SurroFlow operates at a lower resolution, we include a zoomed-in view of the ground truth (bottom row) to ensure a fair visual comparison.}
    \label{fig:data-generation}
    \vspace{-15pt}
\end{figure}

\Cref{fig:data-generation} qualitatively compares the volume-rendered outputs of ground truth and model predictions for two representative datasets: MPAS-O and CloverLeaf3D. 
For each dataset, we visualize one representative example.
In each example, the top row presents the ground truth output alongside predictions from the three models. 
The bottom row displays error maps that highlight voxel-wise differences between predictions and ground truth.
As shown, {\sysname} consistently yields lower reconstruction errors, visually confirming its superior predictive accuracy compared to both VDL and SurroFlow.
These results demonstrate {\sysname}’s ability to deliver accurate and visually faithful surrogate predictions, while also supporting uncertainty quantification and high-resolution outputs. This combination of precision, scalability, and flexibility makes {\sysname} a compelling surrogate modeling solution for complex scientific simulations.

\subsection{Uncertainty Quantification}
In this section, we demonstrate our model’s capability to disentangle aleatoric and epistemic uncertainty, and to accurately capture their respective trends.

\subsubsection{Epistemic uncertainty}

\begin{figure}
    \centering
    \includegraphics[width=\columnwidth]{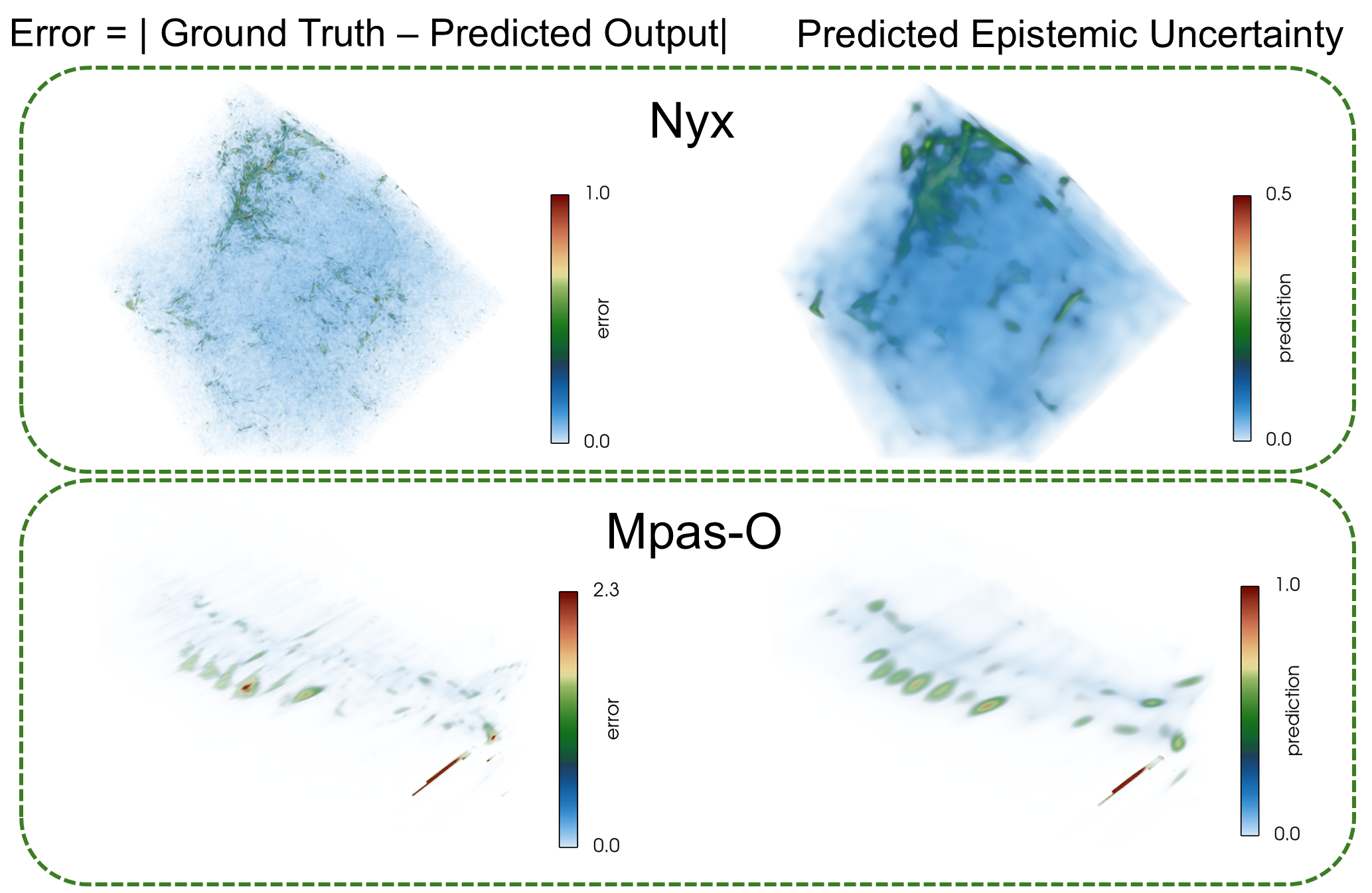}
        \caption{Volume rendering of NYX and MPAS-Ocean data for error maps between ground truth and predicted output (left)  and predicted epistemic uncertainty (right).}
    \label{fig:model-uncertainty}
    \vspace{-2pt}
\end{figure}

Epistemic uncertainty arises from the surrogate model’s lack of knowledge and often correlates with prediction errors. Accordingly, we approximate it using the absolute difference between the predicted output and the ground truth.

\Cref{fig:model-uncertainty} shows a qualitative comparison between the actual prediction error (left) and the predicted epistemic uncertainty (right) for the Nyx and MPAS-O datasets. In MPAS-O, the prediction errors are spatially concentrated, and our model successfully captures both the location and structure of high-error regions. In Nyx, errors are more scattered, but areas with higher uncertainty still align well with regions of larger error. These results demonstrate that {\sysname} effectively identifies regions of low confidence, with its predicted uncertainty closely matching the actual error patterns across diverse data distributions. This indicates strong potential for reliable uncertainty-aware surrogate modeling.


\begin{table}[tb]
\small
\centering
\caption{Testing time, PSNR, Voxel-level Correlation, and Member-level Correlation for {\sysname}'s epistemic uncertainty prediction.}
\label{tab:table-4}
\vspace{-2pt}
\begin{tabular}{c|c|c|c|c|c}
Dataset & Method & $\downarrow${\makecell{Testing \\Time}} & $\uparrow$PSNR & $\uparrow${\makecell{Voxel-level\\ Corr}} & $\uparrow${\makecell{Member-level \\Corr}} \\ 
\hline
\multirow{3}*{\makecell{MPAS-\\Ocean}} & MCD & 2.88s & 34.92 & 0.4252 & 0.5764 \\
  ~ & DE & 1.60s & 28.58 & 0.1050 & 0.2747 \\
  ~ & Ours & \textbf{0.10s} & \textbf{49.73} & \textbf{0.7311} & \textbf{0.5990}\\
\hline
\multirow{3}*{Nyx} & MCD & 25.17s & 28.05 & 0.1260 & 0.7043 \\
  ~ & DE & 2.84s & 29.58 & 0.0416 &  -0.3887 \\
  ~ & Ours & \textbf{0.36s} & \textbf{37.09} & \textbf{0.3105} & \textbf{0.9957}\\
\hline
\multirow{3}*{\makecell{Clover-\\leaf3D}} & MCD & 0.83s & 47.71 & 0.1238 & 0.4629 \\
  ~ & DE & 3.00s & 47.52 & 0.0566 &  0.5396\\
  ~ & Ours & \textbf{0.03s} & \textbf{47.73} & \textbf{0.2428} & \textbf{0.9321} \\
\end{tabular}
\vspace{-12pt}
\end{table}

Based on our quantitative analysis, we compare {\sysname} against two widely used methods for estimating epistemic uncertainty:
\begin{itemize}
  \item Monte Carlo Dropout (shorted for MCD)~\cite{gal2016dropout}: applies dropout during both training and inference to generate multiple predictions, estimating uncertainty from their variance.
  \item Deep Ensemble (shorted for DE)~\cite{lakshminarayanan2017simple}: trains multiple independently initialized models and estimates uncertainty based on the variance of their outputs.
\end{itemize}
To ensure a fair comparison, both baselines are implemented using the same model architecture described in~\cref{sect:Model Architecture} and trained with the MSE loss. 
For MCD, we set dropout rates to 0.2 (fully connected) and 0.3 (convolutional layers). DE consists of five models with different random initializations.

We evaluate prediction accuracy using PSNR and the quality of uncertainty estimation using the Pearson correlation coefficient (Corr) between predicted uncertainty and actual error. Corr values range from \(-1\) (strong negative correlation) to \(1\) (strong positive correlation), with higher values indicating better alignment between predicted uncertainty and true error.

To comprehensively evaluate the alignment between predicted uncertainty and true error, we consider two levels of correlation: \textbf{fine-grained voxel-level correlation} and \textbf{global member-level correlation}.
First, we compute a \textbf{voxel-level correlation} to assess how well the model captures spatial error patterns within individual ensemble members.
Each ensemble member is a high-dimensional volume that may contain spatially localized high-uncertainty regions, making it essential to evaluate the accuracy of the model's uncertainty predictions at a local level.
Second, we calculate a \textbf{member-level correlation} to capture variation in prediction difficulty across ensemble members.
Some ensemble members may be easier to predict with consistently low errors, while others are inherently more challenging, resulting in uniformly higher errors. 
Capturing such member-level variation helps identify difficult cases and improves model robustness.
To formally define these correlations, consider a test dataset comprising \( M \) ensemble members, each containing \( N \) voxels. For voxel \( i \) within ensemble member \( m \), we denote the predicted uncertainty as \( u_{i}^{m} \) and the absolute prediction error as \( e_{i}^{m} \). 
The correlations are computed as follows:

\begin{itemize}
  \item \textbf{Voxel-level correlation} (fine-grained assessment):
  \vspace{-8pt}
  \[
    \text{Corr}_{\text{voxel}} = \frac{1}{M}\sum_{m=1}^{M}\text{Corr}\left(\mathbf{u}^{m}, \mathbf{e}^{m}\right),
  \]\vspace{-4pt}
    \vspace{-4pt}
  \[\text{where} \quad
    \mathbf{u}^{m} = [u_{1}^{m}, u_{2}^{m}, \dots, u_{N}^{m}], \quad 
    \mathbf{e}^{m} = [e_{1}^{m}, e_{2}^{m}, \dots, e_{N}^{m}].
  \]

  \item \textbf{Member-level correlation} (global assessment): 
  \vspace{-6pt}
  \[
    \text{Corr}_{\text{member}} = \text{Corr}\left(\bar{\mathbf{u}}, \bar{\mathbf{e}}\right),
  \]\vspace{-8pt}
  \vspace{-2pt}
  \[\text{where} \quad
  \bar{u}^{m}=\frac{1}{N}\sum_{i=1}^{N}u_{i}^{m}, \quad \bar{e}^{m}=\frac{1}{N}\sum_{i=1}^{N}e_{i}^{m},
  \]
  \[\text{and} \quad
    \bar{\mathbf{u}} = [\bar{u}^{1}, \bar{u}^{2}, \dots, \bar{u}^{M}], \quad 
    \bar{\mathbf{e}} = [\bar{e}^{1}, \bar{e}^{2}, \dots, \bar{e}^{M}].
  \]
\end{itemize}

\Cref{tab:table-4} compares {\sysname} with two widely used uncertainty estimation methods, MCD and DE.  
Unlike these sampling-based approaches that require multiple forward runs, {\sysname} estimates uncertainty with a single forward pass. This leads to significantly faster inference time, making it highly efficient and suitable for real-time and large-scale applications.
In terms of uncertainty estimation quality, {\sysname} consistently achieves substantially the highest voxel-level and member-level correlation scores. This indicates that its predicted epistemic uncertainty closely aligns with the actual prediction error, making it a reliable measure of model confidence.
At the voxel level, {\sysname} successfully identifies local regions within each volume where predictions are less reliable, offering fine-grained insights into spatial uncertainty.  
At the ensemble member level, our model successfully distinguishes test members with higher overall error, often caused by insufficient training data in certain input regions. This capability provides actionable guidance for targeted data collection and further model improvement.
Moreover, {\sysname} also yields the most accurate predictions, as reflected by its consistently highest PSNR across all datasets. While high accuracy is desirable, it poses additional challenges for uncertainty estimation—smaller and fewer errors leave less signal for learning where predictions may be uncertain. Despite this, {\sysname} maintains a strong correlation with true error patterns, which highlights our model’s robustness.


\subsubsection{Aleatoric uncertainty}
Aleatoric uncertainty arises from the inherent randomness in data generation or simulation processes. Because it is intrinsic to the process, its exact ground truth is fundamentally unknowable.
In this study, we introduce two proxies to approximate distinct sources of aleatoric uncertainty in ensemble datasets.
We also compare our {\sysname} with SurroFlow~\cite{shen2024surroflow}, a flow-based surrogate model that captures aleatoric uncertainty by learning complex output distributions from ensemble data.

A major contributor to aleatoric uncertainty in scientific simulations is limited spatial resolution. 
High-resolution volumes preserve fine-grained structures and high-frequency details, while downsampling tends to blur or remove these features in low-resolution versions. This leads to multiple plausible low-resolution realizations for the same high-resolution ground truth, particularly in complex regions. As a result, aleatoric uncertainty is typically higher in structurally rich areas and lower in smoother regions.
In practice, low-resolution data is more accessible due to storage and computational constraints, while high-resolution simulations are expensive but more informative. By accurately identifying high-uncertainty regions within low-resolution volumes, we can selectively generate localized high-resolution samples. This enables more efficient ensemble construction by focusing resources on the most uncertain or information-rich areas.

To evaluate whether {\sysname} can effectively capture resolution-induced aleatoric uncertainty, we train and test the model on low-resolution data derived from high-resolution simulations, and compare its performance with the baseline SurroFlow.
Specifically, we construct a reference "ground truth" uncertainty by measuring the variability across multiple downsampled versions of each high-resolution member and compare this with the aleatoric uncertainty predicted by {\sysname}.
To generate these variants, we apply four downsampling methods: trilinear interpolation, nearest-neighbor interpolation, max pooling, and min pooling. We conduct two experiments: (1) the model is trained with access to all downsampled versions of each member; and (2) the model is trained with only one randomly selected downsampling variant per member. The latter simulates a more realistic scenario in which uncertainty naturally arises due to limited resolution.

\begin{table}[tb]
\small
\centering
\caption{Voxel-level Correlation for aleatoric uncertainty caused by resolution limitations.}
\vspace{-2pt}
\label{tab:table-uncertainty-comparison}
\begin{tabular}{c|c|c|c|c}
Dataset & \makecell{Downsample\\factor} & Method & $\uparrow$\makecell{All \\downsampling\\ results} & $\uparrow$\makecell{One \\random\\ result}\\ 
\hline
\multirow{2}{*}{Nyx} & \multirow{2}{*}{$4\times$} & SurroFlow & 0.2824 & 0.2301\\ 
 &  & Ours & \textbf{0.3625} & \textbf{0.2910} \\
\hline
\multirow{2}{*}{Mpas-O} & \multirow{2}{*}{$2\times$} & SurroFlow & 0.3351 & 0.2713 \\
 &  & Ours & \textbf{0.4254} & \textbf{0.2984} \\
\hline
\multirow{2}{*}{Cloverleaf3D} & \multirow{2}{*}{$4\times$} & SurroFlow & 0.1462 & 0.1121 \\
 &  & Ours & \textbf{0.2503} & \textbf{0.2339} \\
\end{tabular}
\end{table}

\Cref{tab:table-uncertainty-comparison} reports the voxel-level correlation between the predicted aleatoric uncertainty and the reference variability. 
Since this resolution-induced aleatoric uncertainty is inherently spatial—manifesting differently across regions within each member—we report only voxel-level correlations.
Results show that {\sysname} captures the spatial distribution of resolution-induced uncertainty more accurately than the baseline method, SurroFlow, across all three datasets and under both training conditions.
Notably, even when trained with just one random variant per member, our model still achieves relatively good correlation scores. This highlights {\sysname}’s robustness in more practical settings where high-resolution data is scarce and only limited-resolution observations are available.

Beyond resolution-induced uncertainty, another key source of aleatoric uncertainty arises from numerical errors inherent in the simulation process, such as truncation and rounding. We simulate this effect by introducing small perturbations to the input parameters, which lead to output variations reflecting the uncertainty introduced by numerical approximations. These small errors are often amplified through the simulation, providing a measurable proxy for this type of uncertainty.

\begin{figure*}[t]
\centering 
\includegraphics[width=\textwidth]{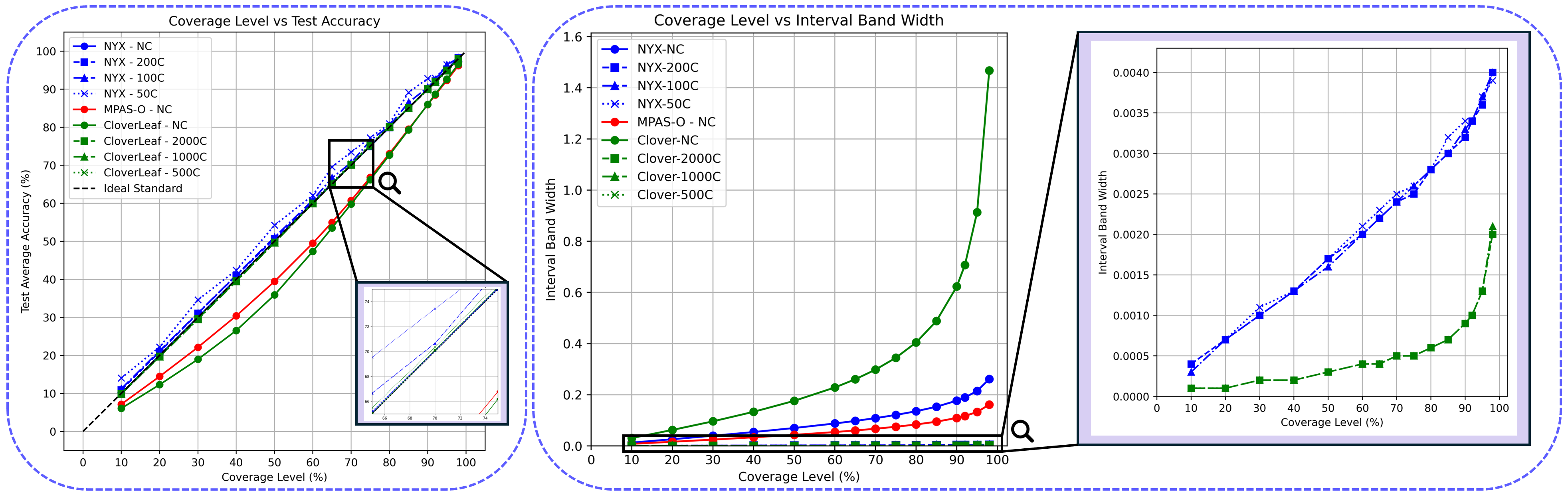}
\vspace{-5pt}
\caption{
Interval prediction results across varying calibration dataset sizes and confidence levels. 
\textit{Left}: Relationship between coverage level and prediction accuracy. 
\textit{Right}: Relationship between coverage level and interval band width. 
Simulation output ranges: Nyx (\([8.77, 12.80]\)), MPAS-O (\([10.09, 29.85]\)), CloverLeaf3D (\([-3.88, 19.69]\)). 
Abbreviations: ``NC'' indicates Non-calibrated (no calibration applied), and numbers followed by ``C'' (e.g., ``50C'', ``200C'') represent the calibration dataset size.
}
\label{fig:interval} 
\vspace{-14pt}
\end{figure*}

\begin{table}[tb]
\small
\centering
\caption{PSNR and Member-level Correlation for aleatoric uncertainty from simulation-induced numerical errors.}
\label{tab:table-6}
\begin{tabular}{c|c|c|c|c}
Dataset & Data Dimension & Method & $\uparrow$PSNR & $\uparrow$Member-level Corr\\ 
\hline
\multirow{2}{*}{Nyx} & [128,128,128] & SurroFlow & 30.9263 & 0.2011 \\
                   & [256,256,256] & Ours & \textbf{37.0429} & \textbf{0.8057} \\
\hline
\multirow{2}{*}{\makecell{Clover-\\leaf3D}} & [64, 64, 64] & SurroFlow & 32.7111 & 0.0155 \\
    & [64, 64, 64] & Ours & \textbf{47.7264} & \textbf{0.5156} \\
\end{tabular}
\vspace{-12pt}
\end{table}

\Cref{tab:table-6} summarizes the results on the Nyx and Cloverleaf3D datasets, reporting the member-level correlation between the predicted aleatoric uncertainty and the reference uncertainty. We focus on member-level correlation here because voxel-level of this ground truth manifests primarily as salt-and-pepper noise, offering limited interpretability. In contrast, the magnitude of numerical errors varies meaningfully across different members, and our {\sysname} successfully captures these trends. Compared to SurroFlow, our model achieves higher correlation scores, indicating more accurate estimation of this source aleatoric uncertainty. This demonstrates the model’s ability to identify member-level uncertainty caused by numerical instability, offering scientists valuable insights into which simulations are less numerically reliable.

Together, the two types of aleatoric uncertainty offer complementary perspectives: the first captures spatial ambiguity due to limited resolution, while the second reflects global variability tied to simulation precision. To further evaluate {\sysname}’s ability to handle inherently stochastic outputs, we inject synthetic noise into the simulation results to mimic randomness. The corresponding experiments and results are provided in the supplemental material.


\subsection{Interval Prediction}

\begin{table}[tb]
\vspace{-1pt}
\small
\centering
\caption{Calibration data sizes and corresponding calibration times for Nyx and Cloverleaf datasets.}
\vspace{-3pt}
\begin{tabular}{c|c|c}
\textbf{} & Calibration Data Size & Calibration Time \\ \hline
\multirow{3}{*}{Nyx} & 200 & 87min13s \\
 & 100 & 42min9s \\ 
 & 50 & 25min15s \\ \hline
\multirow{2}{*}{Cloverleaf} & 2000 & 15min38s \\
 & 1000 & 6min10s\\ 
 & 500 & 2min42s \\
\end{tabular}
\label{tab:calibration_times}
\vspace{-17pt}
\end{table}


This section evaluates {\sysname}'s ability to generate prediction intervals for simulation outputs. While evidential models inherently produce prediction intervals as part of their output, these intervals do not offer formal statistical guarantees—that is, the ground truth may not fall within the predicted range at a specified confidence level.
To address this limitation, we apply the conformal prediction step to calibrate the evidential intervals. This technique adjusts the interval bounds to achieve valid coverage guarantees, often resulting in tighter intervals that retain informativeness.
We evaluate the calibration performance using different calibration set sizes. For the Nyx dataset, we try sizes of 200, 100, and 50; for Cloverleaf3D, 2000, 1000, and 500. For MPAS-O, we simulate a more constrained real-world scenario where no calibration is applied due to the limited availability of ensemble data; raw evidential intervals are used instead.

\Cref{fig:interval} presents the performance of predicted intervals in terms of empirical coverage accuracy (left) and interval width (right) across varying confidence levels. For coverage accuracy, an ideal system should align with the diagonal line \(y = x\), meaning that the predicted confidence level matches the actual fraction of ground truth values captured, neither overconfidence nor excessive conservatism. 
As shown, uncalibrated evidential intervals often loosely follow this trend but deviate notably at lower confidence levels. Even so, they still provide more informative and reliable uncertainty estimates than point predictions alone—particularly when ensemble data is limited.
With access to an additional calibration dataset, conformal prediction significantly improves alignment with the ideal calibration line, yielding statistically valid and well-calibrated intervals. In addition to better coverage accuracy, it also produces noticeably narrower interval bands, as shown in the right panel of \cref{fig:interval}. These tighter intervals indicate that the model expresses uncertainty more precisely, offering higher-confidence predictions with greater utility for downstream scientific tasks.
Interestingly, we observe that calibration performance is relatively robust to the size of the calibration set. For instance, in the Nyx dataset, even small calibration sets (e.g., 50 samples) yield intervals that closely match the intended confidence levels with minimal degradation in sharpness. This suggests that conformal calibration remains effective even in data-scarce settings.

\Cref{tab:calibration_times} summarizes the trade-off between calibration set size and computational cost. While larger calibration sets generally lead to improved calibration stability, they incur longer runtimes. Nevertheless, even with limited calibration data, conformal prediction achieves substantial gains in both interval sharpness and reliability—highlighting its practicality and scalability in real-world simulation scenarios.

\section{Conclusion and Future Work}

In this paper, we propose {\sysname}, a novel conformalized evidential surrogate model for uncertainty quantification in scientific simulations. Assuming simulation outputs follow a higher-order Student-t distribution, {\sysname} learns to accurately predict outputs while disentangling epistemic and aleatoric uncertainties and generating informative predictive intervals.
To further enhance the reliability of these intervals, we calibrate these intervals using conformal prediction, yielding rigorous coverage guarantees with tighter bounds. We also develop an interactive visualization interface that enables intuitive exploration of predictions and uncertainties.
Experimental results show that {\sysname} achieves state-of-the-art accuracy, effectively distinguishes uncertainty sources, and produces narrower, more reliable prediction intervals than existing surrogate models.

In the future, there are still several future directions we can explore. First, we can explore alternative, potentially data-driven non-conformity scores to improve the flexibility of predictive intervals. Second, applying {\sysname} to downstream tasks—such as adaptive sampling, risk-aware optimization, or scientific steering—can enhance decision-making. Third, integrating physics-informed models may further strengthen our framework.

\bibliographystyle{abbrv-doi-hyperref}

\bibliography{template}

\end{document}